\documentclass[sigconf,screen=true]{acmart}

\def\BibTeX{{\rm B\kern-.05em{\sc i\kern-.025em b}\kern-.08emT\kern-.1667em\lower.7ex\hbox{E}\kern-.125emX}}
    
\copyrightyear{2019}
\acmYear{2019}
\setcopyright{acmlicensed}
\acmConference[Paris '19]{Paris '19: The 42nd International ACM SIGIR Conference}{June 21--25, 2019}{Paris, France}

\usepackage{subcaption}
\usepackage[skip=0pt]{caption}
\usepackage{xcolor}
\usepackage{multirow}
\usepackage[inline]{enumitem}

\begin{document}

%
\title[Do Attention Heads Provide Transparency?]{Do Transformer Attention Heads Provide Transparency\\ in Abstractive Summarization?}

%

\author{
Joris Baan$^{1,~2}$\quad 
Maartje ter Hoeve$^1$\quad 
Marlies van der Wees$^2$\quad 
Anne Schuth$^2$\quad
Maarten de Rijke$^1$
}
\affiliation{%
$^1$University of Amsterdam, Amsterdam\quad
$^2$De Persgroep, Amsterdam\\
joris.baan@student.uva.nl, m.a.terhoeve@uva.nl,  marlies.van.der.wees@persgroep.nl, \\
anne.schuth@persgroep.nl, derijke@uva.nl
}

\renewcommand{\shortauthors}{J. Baan et al.}

%

%
\begin{abstract}
Learning algorithms become more powerful, often at the cost of increased complexity. 
In response, the demand for algorithms to be transparent is growing. 
In NLP tasks, attention distributions learned by attention-based deep learning models are used to gain insights in the models' behavior. 
To which extent is this perspective valid for all NLP tasks? 
We investigate whether distributions calculated by different attention heads in a transformer architecture can be used to improve transparency in the task of abstractive summarization. 
To this end, we present both a qualitative and quantitative analysis to investigate the behavior of the attention heads. 
We show that some attention heads indeed specialize towards syntactically and semantically distinct input. 
We propose an approach to evaluate to which extent the Transformer model relies on specifically learned attention distributions. 
We also discuss what this implies for using attention distributions as a means of transparency. 
\end{abstract}

%
%


%

\maketitle

\section{Introduction}
When trusting a machine-generated summary it may be crucial to have an understanding of how this summary came to be. 
Attention mechanisms~\citep{bahdanau2014neural, luong2015effective} have gained popularity in the context of deep learning-based approaches to summarization~\citep{nallapati2016abstractive, see2017get, vaswani2017attention}. 
Briefly, classic attention mechanisms learn a function that assigns a score to each encoder's hidden state based on its relevancy to the word being decoded. 
Through a weighted average with these softmaxed scores, hidden states with high scores are amplified.
Because they provide an interpretable heatmap over an input sequence, attention mechanisms have been used to gain insights in the behavior of a given model~\citep{lei2017interpretable,choi2016retain,jain2019attention}. 
However, this may be misleading. 
First, in commonly used architectures such as Bidirectional Recurrent Neural Networks (Bi-RNNs)~\cite{schuster1997bidirectional} and Transformers~\cite{vaswani2017attention} a lot of computation takes place between an input token and the hidden state. 
As a result, it is unclear whether the hidden state that an attention weight operates on corresponds to its input token. 
Second, shown heatmaps are usually cherry picked and do not necessarily generalize over all examples \cite{jain2019attention,lei2017interpretable,choi2016retain}. 
Third, different attention distributions can lead to the same model output, which implies   that ``attention is not explanation''~\citep{jain2019attention}.

Can attention distributions in a Transformer model~\citep{vaswani2017attention} trained for abstractive summarization be used to address model transparency for the summarization task? 
Transformer models consist of a modular, multi-headed structure. 
Because of this modularity, we may be able to find distinct interpretable patterns that generalize over a large number of examples. 

We adopt a qualitative and a quantitative approach to investigate the behavior of attention heads. 
For the qualitative approach we visually inspect the encoder self-attention and decoder cross-attention and find that some heads attend to locations, persons, organization nouns, or punctuation. 
We then introduce several metrics to quantitatively evaluate whether the previous findings generalize over 1K news articles as well as different initializations of the model. 
Importantly, by doing so we move away from cherry picked attention heatmaps. 
We then discuss a method to investigate to what extent the Transformer model relies on certain attention distributions, inspired by recent work on adversarial attention~\citep{jain2019attention}.
This raises the question whether adversarial methods invalidate the use of learned attention distribution as a means for transparency.

With this work we contribute: 
\begin{enumerate*}
    \item quantitative metrics that measure the degree to which attention heads specialize towards attending Part-of-Speech (POS), Named Entity (NE) tags and relative position; and 
    \item a method for adversarial attacks on seq2seq Transformers to assess the effect of individual attention heads on model output.
\end{enumerate*}

\section{Related Work}
Transparency in machine learning has become important as models become more complex and more frequently play a role in decision making \cite{gilpin2018explaining, mittelstadt2018explaining}. Terms such as explainability and transparency are hard to define and open for multiple interpretations. 
\citet{gilpin2018explaining} describe an \textit{explanation} to be an answer to ``why questions'' and consider it a trade-off between \textit{interpretability} and \textit{completeness}. \textit{Interpretability} means being understandable to humans, whereas \textit{completeness} covers how well the explanation is faithful to the actual model mechanics. 
\citet{doshi2017towards} note that interpretability can be used to evaluate desiderata besides performance such as causality or trust. 
\citet{mittelstadt2018explaining} argue that \textit{transparency} addresses how a model functions internally. Such a model or its components can be called \textit{transparent} when they can be comprehended entirely.  
Following \cite{doshi2017towards,mittelstadt2018explaining}, we maintain that a fully transparent model should be understandable for a user. 
However, since fully transparent models are not always capable of competitive performance, we argue that a step in the direction of a more interpretable model already provides transparency. 
We want to understand the attention mechanism and its role in the Transformer to assist in the discussion on whether attention provides transparency.

\subsection{Attention for transparency in NLP}
The following recent work is aimed towards a better understanding of attention distributions and whether it can be used to explain a model.
\citet{raganato2018analysis} study the self-attention of a Transformer encoder for NMT and observe that some heads mark syntactic dependency relations.
\citet{vig2018deconstructing} visualizes BERT's \cite{devlin2018bert} self-attention and finds patterns such as attention to the surrounding words, identical/related words, predictive words and delimiter tokens. 
Concurrent to our work, \citet{voita2019analyzing} perform a similar analysis of multi-headed attention in NMT and \citet{michel2019sixteen} for BERT \cite{devlin2018bert}. They find that heads specialize towards linguistically interpretable roles, but that a majority can be pruned after training without affecting performance.
\citet{jain2019attention} observe that attention is commonly (implicitly or explicitly) claimed to provide insight into model dynamics. 
They argue that if attention is used as explanation, it should exhibit two properties: 
\begin{enumerate*}
\item attention should correlate with feature importance measures; and 
\item adversarially crafted attention distributions should lead to different predictions, or be considered equally plausible explanations.
\end{enumerate*}
Such an adversarial attention distribution should maximally differ from the learned attention, while the corresponding output distribution is constrained to be the same within a small range $\epsilon$.
With a Bi-RNN or CNN encoder it is possible to construct such adversarial distributions for NLP classification tasks such as binary text classification. They argue that attention heatmaps should thus not be so easily assumed to provide transparency for model predictions~\cite{jain2019attention}.

\subsection{The Transformer}
The Transformer is a seq2seq model that relies solely on \mbox{(self-)at}\-ten\-tion and stacks several encoders and decoders. 
\emph{Self-attention} computes scores between each of the input tokens, as opposed to computing scores between encoder and decoder hidden states, referred to as \emph{cross-attention}. 
\textit{Multi-headed} attention refers to having multiple ``representation subspaces''  or \emph{heads} governed by separate sets of $W_Q, W_K, W_V$ weight matrices. These matrices project each input into a query, key and value vector from which scores and context vectors are computed. 
The attention function itself is \textit{scaled dot product attention} and identical to \citet{luong2015effective}'s dot-product attention apart from the scaling factor (Eq.~(\ref{eq:transf3})). $H$ represents an embedding for the bottom encoder/decoder and a hidden state for the remainder (Eq.~(\ref{eq:transf2})).
\begin{align}
    \label{eq:transf2}
    \mathit{head}_i &= \mathit{Attention}(HW_Q^i, HW^i_K,HW^i_V)\\
    \label{eq:transf3}
    \mathit{Attention}(Q,K,V) &= \mathrm{softmax}\left(\frac{QK^T}{\sqrt{d_k}V}\right).
\end{align}
Our work extends and differs from the related work just discussed in three important ways: \begin{enumerate*}
    \item we introduce three metrics relevant to summarization to quantify patterns in attention;
    \item we analyze the decoder cross-attention in addition to the encoder self-attention; and
    \item our input sequences (news articles) are significantly longer than the short sentences used in previous work.
\end{enumerate*}

\section{Experimental Setup}
\label{section:experimental-setup}
We adopt OpenNMT's implementation \cite{2017opennmt} of the CopyGenerator Transformer~\citep{gehrmann2018bottom}. 
Both encoder and decoder have four layers with eight heads. 
We use scaled dot attention, \citet{gehrmann2018bottom}'s new summary specific coverage function, \citet{wu2016google}'s length penalty during beam-search decoding at inference time, and \citet{see2017get}'s pointer generator architecture. 

We use the \textit{CNN/Daily Mail} \citep{hermann2015teaching,nallapati2016abstractive} dataset containing roughly 300.000 news articles and use the script provided by \citet{nallapati2016abstractive} to split this into a train, test and validation set. 
Articles consist on average of 781 tokens and summaries of 56 tokens. 
Following \citet{see2017get} we truncate articles to 400 words. 
We train two identical models with different parameter initializations to investigate whether stochasticity affects the way attention heads specialize. 
Both models have similar ROUGE scores (\textbf{ROUGE-1}: 38.76/38.81, \textbf{ROUGE-2}: 17.13/16.77, \textbf{ROUGE-L}: 36.00/36.28).

\section{Qualitative Approach}

We extend a tool originally created to visualize a copy-generator model by \citet{see2017get}. 
It highlights words in an input article based on the magnitude of their corresponding attention weights and gives control over which attention type, layer or head to visualize.\footnote{Our version is publicly available at \url{https://aijoris.github.io/attnvis/}.} We compute an overall attention distribution by summing and normalizing attention weights over all time steps.

For the encoder, the vast majority of the attention heads seem to focus on preceding, succeeding or surrounding words. Similarly for the decoder, several heads find an occurrence of the currently or previously decoded word. Some heads seem to focus on punctuation and delimiters overall, confirming observations from \citet{vig2018deconstructing}. 

\begin{figure}[t]
  \centering
  \includegraphics[width=\linewidth]{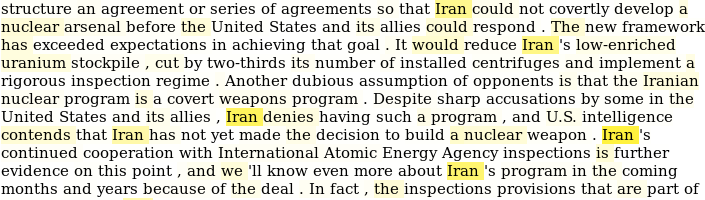}
  \caption{Attention head focusing on locations.}
  \label{fig:locations}
  \vspace{-0.5\baselineskip}
\end{figure}

\begin{figure}[t]
  \centering
  \includegraphics[width=\linewidth]{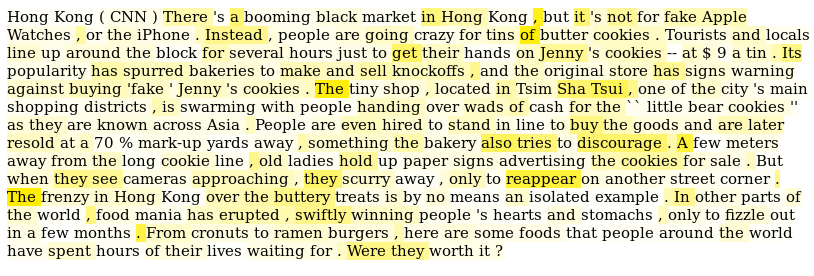}
  \caption{Attention head that seemed to focus on named entities fails to do so in the above example.}
  \label{fig:counter_example}
  \vspace{-0.5\baselineskip}
\end{figure}

Strikingly, when inspecting the overall decoder attention, there are heads that seem to focus on key words, locations (Figure~\ref{fig:locations}), organizations, people or days of the week. 
These heads appear to have learned to detect such entities without explicit supervisory signals.
However, there are plenty of articles for which these patterns are less obvious (Figure~\ref{fig:counter_example}). 
Such ``counter examples'' might indicate that these patterns do not generalize and are based on our bias for interpretability, or the model might sometimes fail to predict the specialized attention, similar to how the ROUGE score is lower for some documents than others.


\section{Quantitative approach}
To support our findings from the qualitative visualizations and examine to what extent the observation generalize, we introduce three quantitative metrics. 

\subsection{Relative position}
\begin{figure}[h]
  \centering
  \includegraphics[width=0.45\linewidth]{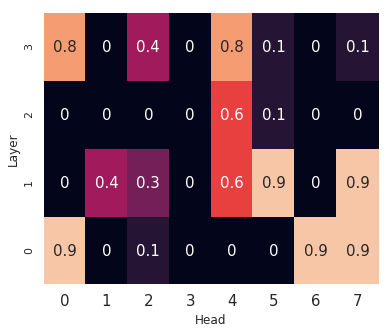}
  \includegraphics[width=0.45\linewidth]{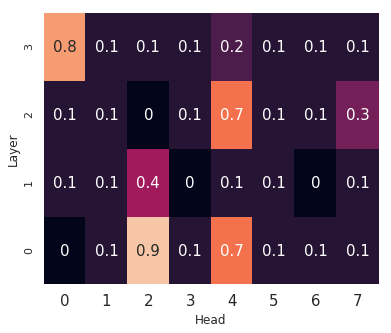}
  \includegraphics[width=0.067\linewidth]{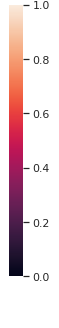}
  \caption{Ratio of the max attention weight being assigned to neighboring tokens. (Left): Encoder. (Right): Decoder.}
  \label{fig:loc_heatmap}
  \vspace{-0.5\baselineskip}
\end{figure}

We record how often the maximum attention weight is on preceding or successive tokens relative to the token currently being encoded or decoded. 
Figure~\ref{fig:loc_heatmap} shows that at least nine encoder heads and four decoder heads focus on relative positions. 
This behavior brings to mind the inductive bias in an RNN where tokens are explicitly processed sequentially, or a CNN using convolutions to construct hidden states. 
The Transformer does not have such inductive bias and solely uses attention.
Interestingly, some heads appear to have learned a similar way of processing nonetheless. 
Model 2 is not shown, but has
a similar number of relative position heads.

\begin{figure*}[t]
  \centering
  \begin{subfigure}[b]{0.485\textwidth}
  \includegraphics[width=\textwidth]{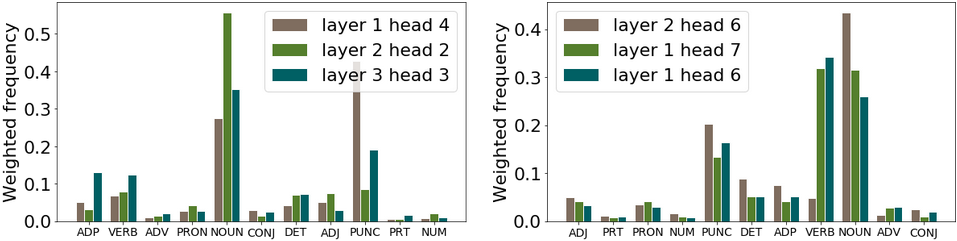}
  \caption{Based on POS-KL. (Left): Model 1; (Right): Model 2.}
  \label{fig:pos_plot}
  \end{subfigure}
  \quad
  \begin{subfigure}[b]{0.485\textwidth}
  \includegraphics[width=\textwidth]{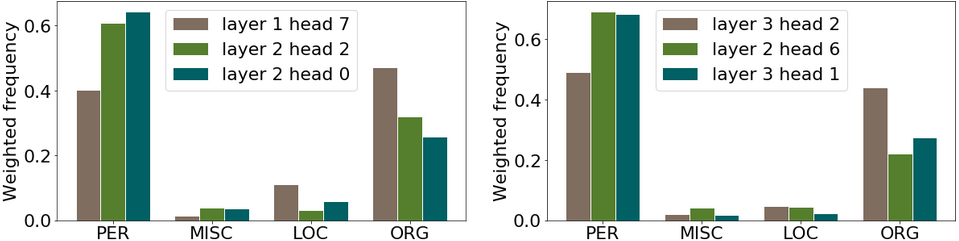}
  \caption{Based on NEP. (Left): Model 1; (Right): Model 2.}
  \label{fig:ner_plot}
  \end{subfigure}
  \caption{A comparison of the top 3 specialized heads.}
  \vspace{-\baselineskip}
\end{figure*}

\subsection{POS-KL}
We tag each article in the test set (see Section~\ref{section:experimental-setup}) with $12$ universal part-of-speech tags \cite{petrov2011universal} using a POS tagger by FLAIR \cite{akbik2018coling}.
For every article we compute a histogram of POS tag counts to serve as the baseline.
Then, for each head these tag counts are multiplied by the accumulated attention weights of all tokens labeled with that tag and normalized.
The degree to which an attention head is specialized can be measured by the difference between its attention-weighted POS tag distribution and the baseline POS tag distribution.
We use the KL-divergence to quantify this difference and average these over all articles.

Figure~\ref{fig:pos_plot} shows the decoder cross-attention weighted POS tag distributions for three heads with the highest KL-divergence. 
The peaks at the punctuation, noun and verb tag confirm that some heads consistently focus on specific word categories.
For the two trained models, different specializations emerge.
Model~2 has two heads with a large peak for verbs, and all three heads have a relatively high peak at punctuation as well.
Model~1 has no such peaks for verbs and only one of the top three heads that focuses on punctuation. 

\subsection{NEP}
Each article is tagged with four named entities: \textit{persons}, \textit{locations}, \textit{organizations}, and \textit{miscellaneous}, using a NE tagger by FLAIR \cite{akbik2018coling}. Unlike POS tags, however, not each token is a named entity. This can cause a high KL-divergence between the attention weighted named entity distribution and baseline (NE-KL), even if a head barely attends to named entities. We found computing the proportion of attention mass over all named entities (NEP) to be a better method for detecting specialized heads. 

The baseline ratio of named entities over articles is 0.1. Figure~\ref{fig:ner_plot} shows the top three cross-attention weighted distributions over named entities based on NEP. Heads shown have a NEP of at least double the baseline ratio. Large peaks at persons and organizations can be observed for both models. Model 1's most specialized head corresponds to the `location head' found in our qualitative analysis. This indicates the ability to detect specialized heads using NEP. It simultaneously provides more insight into what such a head actually attends and how well our qualitative findings generalize. We refer the reader to the appendix for a complete overview of the metrics for all attention heads, including standard deviations.

\subsection{Analysis}
We did not detect any POS or NE specialization for the encoder's overall self-attention.
This is in line with the earlier observation that most encoder heads attend relative positions. 
It is important to note that we have not evaluated the models on per-document ROUGE scores. 
This could explain the observed difference in specialization between models.
Perhaps model 2 performs better on articles for which verbs are important in the summary, resulting in a head that more explicitly attends verbs.
Another note is that not every article contains named entities, causing a decrease in NEP. One interesting example can be found in Appendix A, where a NE-specialized head highlights lions in one article. Lions are not named entities but do fulfil a similar role, indicating that NEP might not always fully capture a specialization.

The main takeaway is that we show that some attention heads specialize towards attending relative locations, nouns, verbs, punctuation, persons, locations or named entities. 
The top 3 specialized heads that were found using our quantitative approach line up with findings from visualizations. 
However, an analysis of POS-KL and NEP distributions over articles also indicate that heads only specialize to some extend and sometimes take into account a considerable amount of non-related tokens. 
This supports claims by \citet{jain2019attention}, urging the research community to be careful in using attention as explanation.

\section{Adversarial Attention}
Given that some attention heads are found to focus on interpretable input, we want to understand to what extent the model actually relies on these specific attention distributions.
For future work, we propose to adapt the adversarial attention method by \cite{jain2019attention} to make it compatible with a seq2seq Transformer model using beam search.
Instead of requiring the output distributions to be within a small $\epsilon$, it is sufficient to constrain the top $K$ output probabilities of each decoding to be within a small $\epsilon$, whereby $K=beamsize$.
This will result in identical output sequences, since the beam search path with the highest probability remains the same. As a consequence, we can craft one adversarial attention distribution for each decoding step and aggregate them to evaluate the overall success on a summary.

Additionally, we propose to modify the attention distribution of a specialized head to attend another specific phenomenon. 
For example, we could construct a distribution that solely attends persons for a head that specializes towards locations and observe whether locations change into persons. 
The Transformer model is large, in our case containing 32 heads for both the encoder and decoder.
It is unclear to what degree modifying the attention distribution of one head can be expected to affect the output summary. 

However, if such an adversarial distribution can be constructed, it raises the question to what extend it invalidates the learned attention distribution as means for transparency.
Should an attention distribution have a causal relationship with the model output in order to use it for transparency, or does the fact that the model has learned this distribution justify using it as such?
Similarly, does the use of attention heads to address transparency become invalidated if different specializations form for architecturally identical model on the same data set? Or does this add to its value because it shows differences between models that otherwise remain undetected?

\section{Conclusion}
We have presented a qualitative and quantitative approach to better understand what Transformer attention heads attend to in abstractive summarization. 
Some attention heads do specialize towards interpretable parts of a document, but this does not apply to all documents.
We confirm this with three proposed metrics that quantify what heads focus on in terms of POS tags, named entities and relative position.
We also find that these specializations are not consistent over differently initialized models.
Finally, we discuss the use of adversarial attention to examine the effect of attention distributions on model output, and ask what such adversarial methods  imply for transparency. 

One limitation of this work is that there is no proof that the index of a hidden state corresponds to a (contextual) representation of the corresponding input token.
A natural question is why specialized heads perform poorly on some articles. Future work could compare per-document ROUGE with POS-KL and NEP to examine correlations between summarization and head specialization.


\begin{acks}
We thank Mostafa Dehghani for valuable discussions and feedback.
This research was partially supported by 
Ahold Delhaize,
the Association of Universities in the Netherlands (VSNU),
the Innovation Center for Artificial Intelligence (ICAI),
and the Police AI lab.
All content represents the opinion of the authors, which is not necessarily shared or endorsed by their respective employers and/or sponsors.
\end{acks}

\bibliographystyle{ACM-Reference-Format}
\bibliography{acmart.bib}


\begin{thebibliography}{23}


\ifx \showCODEN    \undefined \def \showCODEN     #1{\unskip}     \fi
\ifx \showDOI      \undefined \def \showDOI       #1{#1}\fi
\ifx \showISBNx    \undefined \def \showISBNx     #1{\unskip}     \fi
\ifx \showISBNxiii \undefined \def \showISBNxiii  #1{\unskip}     \fi
\ifx \showISSN     \undefined \def \showISSN      #1{\unskip}     \fi
\ifx \showLCCN     \undefined \def \showLCCN      #1{\unskip}     \fi
\ifx \shownote     \undefined \def \shownote      #1{#1}          \fi
\ifx \showarticletitle \undefined \def \showarticletitle #1{#1}   \fi
\ifx \showURL      \undefined \def \showURL       {\relax}        \fi
\providecommand\bibfield[2]{#2}
\providecommand\bibinfo[2]{#2}
\providecommand\natexlab[1]{#1}
\providecommand\showeprint[2][]{arXiv:#2}

\bibitem[\protect\citeauthoryear{Akbik, Blythe, and Vollgraf}{Akbik
  et~al\mbox{.}}{2018}]%
        {akbik2018coling}
\bibfield{author}{\bibinfo{person}{Alan Akbik}, \bibinfo{person}{Duncan
  Blythe}, {and} \bibinfo{person}{Roland Vollgraf}.}
  \bibinfo{year}{2018}\natexlab{}.
\newblock \showarticletitle{Contextual String Embeddings for Sequence
  Labeling}. In \bibinfo{booktitle}{\emph{{COLING} 2018, 27th International
  Conference on Computational Linguistics}}. \bibinfo{pages}{1638--1649}.
\newblock


\bibitem[\protect\citeauthoryear{Bahdanau, Cho, and Bengio}{Bahdanau
  et~al\mbox{.}}{2014}]%
        {bahdanau2014neural}
\bibfield{author}{\bibinfo{person}{Dzmitry Bahdanau},
  \bibinfo{person}{Kyunghyun Cho}, {and} \bibinfo{person}{Yoshua Bengio}.}
  \bibinfo{year}{2014}\natexlab{}.
\newblock \showarticletitle{Neural Machine Translation by Jointly Learning to
  Align and Translate}.
\newblock \bibinfo{journal}{\emph{arXiv preprint arXiv:1409.0473}}
  (\bibinfo{year}{2014}).
\newblock


\bibitem[\protect\citeauthoryear{Choi, Bahadori, Sun, Kulas, Schuetz, and
  Stewart}{Choi et~al\mbox{.}}{2016}]%
        {choi2016retain}
\bibfield{author}{\bibinfo{person}{Edward Choi}, \bibinfo{person}{Mohammad~Taha
  Bahadori}, \bibinfo{person}{Jimeng Sun}, \bibinfo{person}{Joshua Kulas},
  \bibinfo{person}{Andy Schuetz}, {and} \bibinfo{person}{Walter Stewart}.}
  \bibinfo{year}{2016}\natexlab{}.
\newblock \showarticletitle{Retain: An Interpretable Predictive Model for
  Healthcare using Reverse Time Attention Mechanism}. In
  \bibinfo{booktitle}{\emph{Advances in Neural Information Processing
  Systems}}. \bibinfo{pages}{3504--3512}.
\newblock


\bibitem[\protect\citeauthoryear{Devlin, Chang, Lee, and Toutanova}{Devlin
  et~al\mbox{.}}{2018}]%
        {devlin2018bert}
\bibfield{author}{\bibinfo{person}{Jacob Devlin}, \bibinfo{person}{Ming-Wei
  Chang}, \bibinfo{person}{Kenton Lee}, {and} \bibinfo{person}{Kristina
  Toutanova}.} \bibinfo{year}{2018}\natexlab{}.
\newblock \showarticletitle{Bert: Pre-training of Deep Bidirectional
  Transformers for Language Understanding}.
\newblock \bibinfo{journal}{\emph{arXiv preprint arXiv:1810.04805}}
  (\bibinfo{year}{2018}).
\newblock


\bibitem[\protect\citeauthoryear{Doshi-Velez and Kim}{Doshi-Velez and
  Kim}{2017}]%
        {doshi2017towards}
\bibfield{author}{\bibinfo{person}{Finale Doshi-Velez} {and}
  \bibinfo{person}{Been Kim}.} \bibinfo{year}{2017}\natexlab{}.
\newblock \showarticletitle{Towards a Rigorous Science of Interpretable Machine
  Learning}.
\newblock \bibinfo{journal}{\emph{arXiv preprint arXiv:1702.08608}}
  (\bibinfo{year}{2017}).
\newblock


\bibitem[\protect\citeauthoryear{Gehrmann, Deng, and Rush}{Gehrmann
  et~al\mbox{.}}{2018}]%
        {gehrmann2018bottom}
\bibfield{author}{\bibinfo{person}{Sebastian Gehrmann},
  \bibinfo{person}{Yuntian Deng}, {and} \bibinfo{person}{Alexander~M Rush}.}
  \bibinfo{year}{2018}\natexlab{}.
\newblock \showarticletitle{Bottom-up Abstractive Summarization}.
\newblock \bibinfo{journal}{\emph{arXiv preprint arXiv:1808.10792}}
  (\bibinfo{year}{2018}).
\newblock


\bibitem[\protect\citeauthoryear{Gilpin, Bau, Yuan, Bajwa, Specter, and
  Kagal}{Gilpin et~al\mbox{.}}{2018}]%
        {gilpin2018explaining}
\bibfield{author}{\bibinfo{person}{Leilani~H Gilpin}, \bibinfo{person}{David
  Bau}, \bibinfo{person}{Ben~Z Yuan}, \bibinfo{person}{Ayesha Bajwa},
  \bibinfo{person}{Michael Specter}, {and} \bibinfo{person}{Lalana Kagal}.}
  \bibinfo{year}{2018}\natexlab{}.
\newblock \showarticletitle{Explaining Explanations: An Approach to Evaluating
  Interpretability of Machine Learning}.
\newblock \bibinfo{journal}{\emph{arXiv preprint arXiv:1806.00069}}
  (\bibinfo{year}{2018}).
\newblock


\bibitem[\protect\citeauthoryear{Hermann, Kocisky, Grefenstette, Espeholt, Kay,
  Suleyman, and Blunsom}{Hermann et~al\mbox{.}}{2015}]%
        {hermann2015teaching}
\bibfield{author}{\bibinfo{person}{Karl~Moritz Hermann}, \bibinfo{person}{Tomas
  Kocisky}, \bibinfo{person}{Edward Grefenstette}, \bibinfo{person}{Lasse
  Espeholt}, \bibinfo{person}{Will Kay}, \bibinfo{person}{Mustafa Suleyman},
  {and} \bibinfo{person}{Phil Blunsom}.} \bibinfo{year}{2015}\natexlab{}.
\newblock \showarticletitle{Teaching Machines to Read and Comprehend}. In
  \bibinfo{booktitle}{\emph{Advances in neural information processing
  systems}}. \bibinfo{pages}{1693--1701}.
\newblock


\bibitem[\protect\citeauthoryear{Jain and Wallace}{Jain and Wallace}{2019}]%
        {jain2019attention}
\bibfield{author}{\bibinfo{person}{Sarthak Jain} {and} \bibinfo{person}{Byron~C
  Wallace}.} \bibinfo{year}{2019}\natexlab{}.
\newblock \showarticletitle{Attention is not Explanation}.
\newblock \bibinfo{journal}{\emph{arXiv preprint arXiv:1902.10186}}
  (\bibinfo{year}{2019}).
\newblock


\bibitem[\protect\citeauthoryear{{Klein}, {Kim}, {Deng}, {Senellart}, and
  {Rush}}{{Klein} et~al\mbox{.}}{2017}]%
        {2017opennmt}
\bibfield{author}{\bibinfo{person}{Guillaume {Klein}}, \bibinfo{person}{Yoon
  {Kim}}, \bibinfo{person}{Yuntian {Deng}}, \bibinfo{person}{Jean {Senellart}},
  {and} \bibinfo{person}{Alexander~M. {Rush}}.}
  \bibinfo{year}{2017}\natexlab{}.
\newblock \showarticletitle{{OpenNMT: Open-Source Toolkit for Neural Machine
  Translation}}.
\newblock \bibinfo{journal}{\emph{arXiv preprint arXiv:1701.02810}}
  (\bibinfo{year}{2017}).
\newblock


\bibitem[\protect\citeauthoryear{Lei}{Lei}{2017}]%
        {lei2017interpretable}
\bibfield{author}{\bibinfo{person}{Tao Lei}.} \bibinfo{year}{2017}\natexlab{}.
\newblock \emph{\bibinfo{title}{Interpretable Neural Models for Natural
  Language Processing}}.
\newblock \bibinfo{thesistype}{Ph.D. Dissertation}.
  \bibinfo{school}{Massachusetts Institute of Technology}.
\newblock


\bibitem[\protect\citeauthoryear{Luong, Pham, and Manning}{Luong
  et~al\mbox{.}}{2015}]%
        {luong2015effective}
\bibfield{author}{\bibinfo{person}{Minh-Thang Luong}, \bibinfo{person}{Hieu
  Pham}, {and} \bibinfo{person}{Christopher~D Manning}.}
  \bibinfo{year}{2015}\natexlab{}.
\newblock \showarticletitle{Effective Approaches to Attention-based Neural
  Machine Translation}.
\newblock \bibinfo{journal}{\emph{arXiv preprint arXiv:1508.04025}}
  (\bibinfo{year}{2015}).
\newblock


\bibitem[\protect\citeauthoryear{Michel, Levy, and Neubig}{Michel
  et~al\mbox{.}}{2019}]%
        {michel2019sixteen}
\bibfield{author}{\bibinfo{person}{Paul Michel}, \bibinfo{person}{Omer Levy},
  {and} \bibinfo{person}{Graham Neubig}.} \bibinfo{year}{2019}\natexlab{}.
\newblock \showarticletitle{Are Sixteen Heads Really Better than One?}
\newblock \bibinfo{journal}{\emph{arXiv preprint arXiv:1905.10650}}
  (\bibinfo{year}{2019}).
\newblock


\bibitem[\protect\citeauthoryear{Mittelstadt, Russell, and Wachter}{Mittelstadt
  et~al\mbox{.}}{2018}]%
        {mittelstadt2018explaining}
\bibfield{author}{\bibinfo{person}{Brent Mittelstadt}, \bibinfo{person}{Chris
  Russell}, {and} \bibinfo{person}{Sandra Wachter}.}
  \bibinfo{year}{2018}\natexlab{}.
\newblock \showarticletitle{Explaining Explanations in AI}.
\newblock \bibinfo{journal}{\emph{arXiv preprint arXiv:1811.01439}}
  (\bibinfo{year}{2018}).
\newblock


\bibitem[\protect\citeauthoryear{Nallapati, Zhou, Gulcehre, Xiang,
  et~al\mbox{.}}{Nallapati et~al\mbox{.}}{2016}]%
        {nallapati2016abstractive}
\bibfield{author}{\bibinfo{person}{Ramesh Nallapati}, \bibinfo{person}{Bowen
  Zhou}, \bibinfo{person}{Caglar Gulcehre}, \bibinfo{person}{Bing Xiang},
  {et~al\mbox{.}}} \bibinfo{year}{2016}\natexlab{}.
\newblock \showarticletitle{Abstractive Text Summarization using
  Sequence-to-sequence RNNs and Beyond}.
\newblock \bibinfo{journal}{\emph{arXiv preprint arXiv:1602.06023}}
  (\bibinfo{year}{2016}).
\newblock


\bibitem[\protect\citeauthoryear{Petrov, Das, and McDonald}{Petrov
  et~al\mbox{.}}{2011}]%
        {petrov2011universal}
\bibfield{author}{\bibinfo{person}{Slav Petrov}, \bibinfo{person}{Dipanjan
  Das}, {and} \bibinfo{person}{Ryan McDonald}.}
  \bibinfo{year}{2011}\natexlab{}.
\newblock \showarticletitle{A Universal Part-of-Speech Tagset}.
\newblock \bibinfo{journal}{\emph{arXiv preprint arXiv:1104.2086}}
  (\bibinfo{year}{2011}).
\newblock


\bibitem[\protect\citeauthoryear{Raganato, Tiedemann, et~al\mbox{.}}{Raganato
  et~al\mbox{.}}{2018}]%
        {raganato2018analysis}
\bibfield{author}{\bibinfo{person}{Alessandro Raganato},
  \bibinfo{person}{J{\"o}rg Tiedemann}, {et~al\mbox{.}}}
  \bibinfo{year}{2018}\natexlab{}.
\newblock \showarticletitle{An Analysis of Encoder Representations in
  Transformer-Based Machine Translation}. In \bibinfo{booktitle}{\emph{2018
  EMNLP Workshop BlackboxNLP: Analyzing and Interpreting Neural Networks for
  NLP}}. ACL.
\newblock


\bibitem[\protect\citeauthoryear{Schuster and Paliwal}{Schuster and
  Paliwal}{1997}]%
        {schuster1997bidirectional}
\bibfield{author}{\bibinfo{person}{Mike Schuster} {and}
  \bibinfo{person}{Kuldip~K Paliwal}.} \bibinfo{year}{1997}\natexlab{}.
\newblock \showarticletitle{Bidirectional Recurrent Neural Networks}.
\newblock \bibinfo{journal}{\emph{IEEE Transactions on Signal Processing}}
  \bibinfo{volume}{45}, \bibinfo{number}{11} (\bibinfo{year}{1997}),
  \bibinfo{pages}{2673--2681}.
\newblock


\bibitem[\protect\citeauthoryear{See, Liu, and Manning}{See
  et~al\mbox{.}}{2017}]%
        {see2017get}
\bibfield{author}{\bibinfo{person}{Abigail See}, \bibinfo{person}{Peter~J Liu},
  {and} \bibinfo{person}{Christopher~D Manning}.}
  \bibinfo{year}{2017}\natexlab{}.
\newblock \showarticletitle{Get to the Point: Summarization with
  Pointer-Generator Networks}.
\newblock \bibinfo{journal}{\emph{arXiv preprint arXiv:1704.04368}}
  (\bibinfo{year}{2017}).
\newblock


\bibitem[\protect\citeauthoryear{Vaswani, Shazeer, Parmar, Uszkoreit, Jones,
  Gomez, Kaiser, and Polosukhin}{Vaswani et~al\mbox{.}}{2017}]%
        {vaswani2017attention}
\bibfield{author}{\bibinfo{person}{Ashish Vaswani}, \bibinfo{person}{Noam
  Shazeer}, \bibinfo{person}{Niki Parmar}, \bibinfo{person}{Jakob Uszkoreit},
  \bibinfo{person}{Llion Jones}, \bibinfo{person}{Aidan~N Gomez},
  \bibinfo{person}{{\L}ukasz Kaiser}, {and} \bibinfo{person}{Illia
  Polosukhin}.} \bibinfo{year}{2017}\natexlab{}.
\newblock \showarticletitle{Attention is All You Need}. In
  \bibinfo{booktitle}{\emph{Advances in Neural Information Processing
  Systems}}. \bibinfo{pages}{5998--6008}.
\newblock


\bibitem[\protect\citeauthoryear{Vig}{Vig}{2018}]%
        {vig2018deconstructing}
\bibfield{author}{\bibinfo{person}{Jesse Vig}.}
  \bibinfo{year}{2018}\natexlab{}.
\newblock \bibinfo{title}{Deconstructing BERT: Distilling 6 Patterns from 100
  Million Parameters}.
\newblock
  \bibinfo{howpublished}{\textit{towardsdatascience.com/deconstructing-bert-distilling-6-patterns-from-100-million-parameters-b49113672f77}}.
\newblock
\newblock
\shownote{Accessed: 2019-04-29.}


\bibitem[\protect\citeauthoryear{Voita, Talbot, Moiseev, Sennrich, and
  Titov}{Voita et~al\mbox{.}}{2019}]%
        {voita2019analyzing}
\bibfield{author}{\bibinfo{person}{Elena Voita}, \bibinfo{person}{David
  Talbot}, \bibinfo{person}{Fedor Moiseev}, \bibinfo{person}{Rico Sennrich},
  {and} \bibinfo{person}{Ivan Titov}.} \bibinfo{year}{2019}\natexlab{}.
\newblock \showarticletitle{Analyzing Multi-Head Self-Attention: Specialized
  Heads Do the Heavy Lifting, the Rest Can Be Pruned}.
\newblock \bibinfo{journal}{\emph{arXiv preprint arXiv:1905.09418}}
  (\bibinfo{year}{2019}).
\newblock


\bibitem[\protect\citeauthoryear{Wu et~al\mbox{.}}{Wu et~al\mbox{.}}{2016}]%
        {wu2016google}
\bibfield{author}{\bibinfo{person}{Yonghui Wu} {et~al\mbox{.}}}
  \bibinfo{year}{2016}\natexlab{}.
\newblock \showarticletitle{Google's Neural Machine Translation System:
  Bridging the Gap between Human and Machine Translation}.
\newblock \bibinfo{journal}{\emph{arXiv preprint arXiv:1609.08144}}
  (\bibinfo{year}{2016}).
\newblock


\end{thebibliography}

\appendix
\section{Appendix}
\begin{figure}[h]
  \centering
  \includegraphics[width=\linewidth]{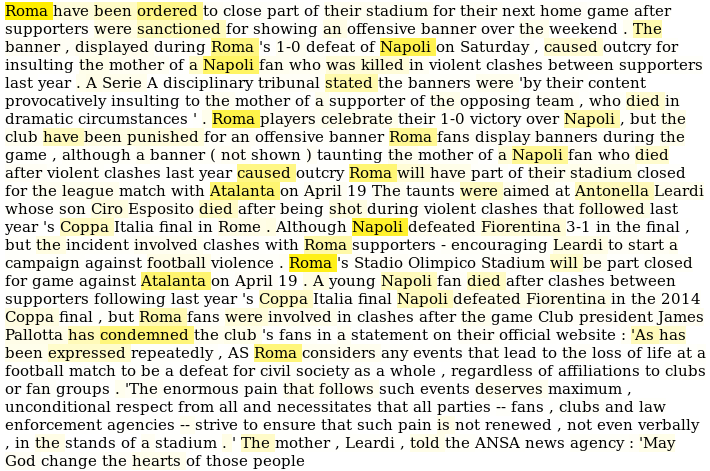}
  \caption{Specialized named entity head focusing on football teams.}
  \label{fig:heatmap1}    \vspace{-0.5\baselineskip}
\end{figure}
\begin{figure}[h]
  \centering
  \includegraphics[width=\linewidth]{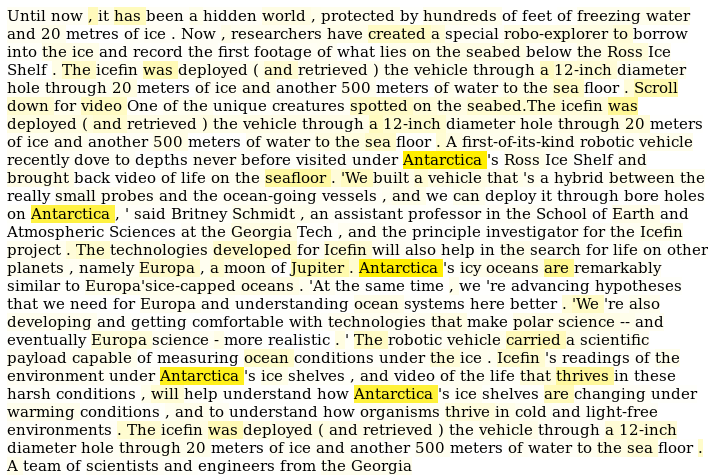}
  \caption{Specialized head focusing on the location Antarctica.}
  \label{fig:heatmap2}
   \vspace{-0.5\baselineskip}
\end{figure}
\vfill\eject
\begin{figure}[h]
  \centering
  \includegraphics[width=\linewidth]{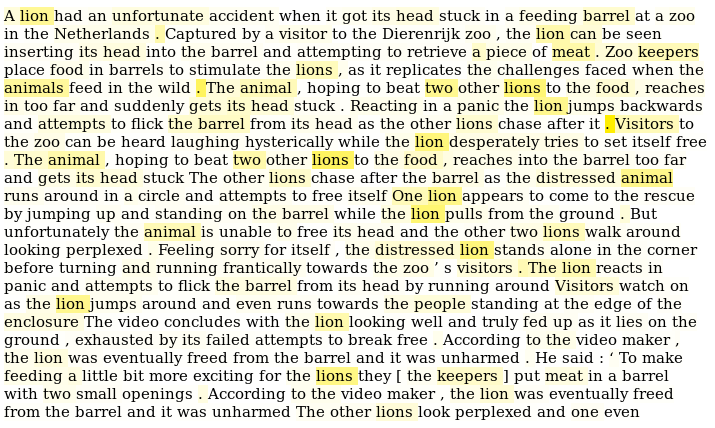}
  \caption{Specialized NE head with a low NEP. This is interesting because this head attends animals in this article, which are not named entities. However, intuitively this example still shows a form of specialization, but this is not reflected by the NEP metric.}
  \label{fig:heatmap3}
  \vspace{-0.5\baselineskip}
\end{figure}
\begin{figure}[h]
  \centering
  \includegraphics[width=\linewidth]{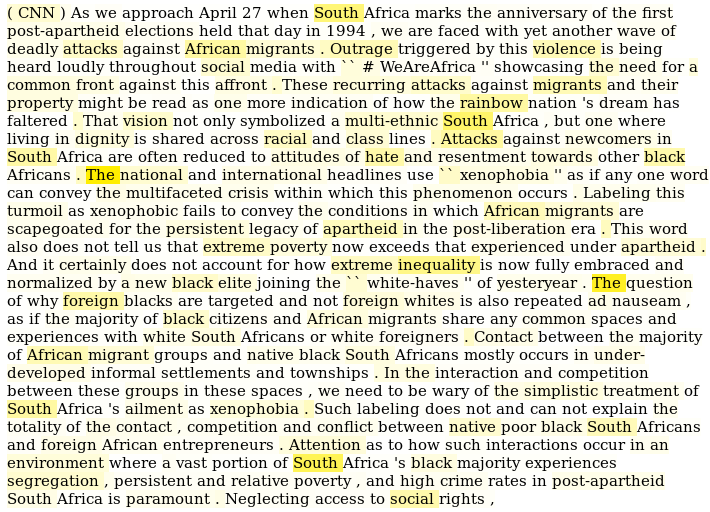}
  \caption{Specialized NE head showing a non interpretable pattern.}
  \label{fig:heatmap4}
   \vspace{-0.5\baselineskip}
\end{figure}

\begin{table*}[]
\caption{Metric scores for the decoder cross attention of model 1. \#1 POS and \#1 NE show the most attended POS tag or named entity for that attention head along with its ratio compared to the other tags. For each column, the three heads with the highest scores are boldfaced.}
\label{tab:model1_crossattn}
\begin{tabular}{ll|l|l|l|l|l|}
\cline{3-7}
                                                         &              & POS-KL & NEP & NER-KL & \#1 POS & \#1 NE \\ \hline
\multicolumn{1}{|l|}{\multirow{8}{*}{Layer 0 }}  	 & Head 0 	& 0.03 $\pm$ \textit{0.02} & 0.15 $\pm$ \textit{0.08} & 0.04 $\pm$ \textit{0.05}& NOUN: 0.340& PER: 0.610\\    \cline{2-7}
\multicolumn{1}{|l|}{}                                   & Head 1 	& 0.05 $\pm$ \textit{0.03} & 0.13 $\pm$ \textit{0.07} & 0.1 $\pm$ \textit{0.09}& NOUN: 0.360& PER: 0.560\\    \cline{2-7}
\multicolumn{1}{|l|}{}                                   & Head 2 	& 0.03 $\pm$ \textit{0.02} & 0.13 $\pm$ \textit{0.08} & 0.06 $\pm$ \textit{0.09}& NOUN: 0.330& PER: 0.490\\    \cline{2-7}
\multicolumn{1}{|l|}{}                                   & Head 3 	& 0.1 $\pm$ \textit{0.04} & 0.1 $\pm$ \textit{0.06} & \textbf{0.21} $\pm$ \textit{0.19}& NOUN: 0.240& PER: \textbf{0.760}\\    \cline{2-7}
\multicolumn{1}{|l|}{}                                   & Head 4 	& 0.04 $\pm$ \textit{0.03} & 0.16 $\pm$ \textit{0.09} & 0.06 $\pm$ \textit{0.05}& NOUN: 0.350& PER: 0.570\\    \cline{2-7}
\multicolumn{1}{|l|}{}                                   & Head 5 	& 0.07 $\pm$ \textit{0.03} & 0.15 $\pm$ \textit{0.08} & 0.15 $\pm$ \textit{0.13}& NOUN: 0.390& PER: 0.630\\    \cline{2-7}
\multicolumn{1}{|l|}{}                                   & Head 6 	& 0.12 $\pm$ \textit{0.05} & 0.09 $\pm$ \textit{0.05} & 0.08 $\pm$ \textit{0.08}& ADP: 0.240& PER: 0.430\\    \cline{2-7}
\multicolumn{1}{|l|}{}                                   & Head 7 	& 0.09 $\pm$ \textit{0.03} & 0.16 $\pm$ \textit{0.07} & 0.1 $\pm$ \textit{0.1}& NOUN: 0.350& PER: 0.520\\    \hline \hline
\multicolumn{1}{|l|}{\multirow{8}{*}{Layer 1 }}  	 & Head 0 	& 0.08 $\pm$ \textit{0.04} & 0.09 $\pm$ \textit{0.06} & 0.12 $\pm$ \textit{0.11}& NOUN: 0.370& PER: 0.520\\    \cline{2-7}
\multicolumn{1}{|l|}{}                                   & Head 1 	& 0.15 $\pm$ \textit{0.06} & 0.17 $\pm$ \textit{0.09} & 0.13 $\pm$ \textit{0.11}& NOUN: 0.300& PER: 0.670\\    \cline{2-7}
\multicolumn{1}{|l|}{}                                   & Head 2 	& 0.15 $\pm$ \textit{0.05} & 0.13 $\pm$ \textit{0.08} & 0.19 $\pm$ \textit{0.15}& NOUN: 0.390& ORG: 0.420\\    \cline{2-7}
\multicolumn{1}{|l|}{}                                   & Head 3 	& 0.07 $\pm$ \textit{0.04} & 0.15 $\pm$ \textit{0.08} & 0.2 $\pm$ \textit{0.17}& NOUN: 0.350& PER: 0.720\\    \cline{2-7}
\multicolumn{1}{|l|}{}                                   & Head 4 	& \textbf{0.42} $\pm$ \textit{0.14} & 0.09 $\pm$ \textit{0.05} & 0.07 $\pm$ \textit{0.07}& PUNC: 0.430& PER: 0.660\\    \cline{2-7}
\multicolumn{1}{|l|}{}                                   & Head 5 	& 0.14 $\pm$ \textit{0.06} & 0.2 $\pm$ \textit{0.09} & 0.11 $\pm$ \textit{0.12}& NOUN: 0.320& PER: 0.640\\    \cline{2-7}
\multicolumn{1}{|l|}{}                                   & Head 6 	& 0.09 $\pm$ \textit{0.06} & 0.15 $\pm$ \textit{0.07} & 0.11 $\pm$ \textit{0.1}& NOUN: 0.350& PER: 0.540\\    \cline{2-7}
\multicolumn{1}{|l|}{}                                   & Head 7 	& 0.13 $\pm$ \textit{0.04} & \textbf{0.27} $\pm$ \textit{0.09} & 0.15 $\pm$ \textit{0.15}& NOUN: 0.380& ORG: 0.470\\    \hline \hline
\multicolumn{1}{|l|}{\multirow{8}{*}{Layer 2 }}  	 & Head 0 	& 0.15 $\pm$ \textit{0.05} & \textbf{0.23} $\pm$ \textit{0.09} & 0.08 $\pm$ \textit{0.1}& NOUN: 0.440& PER: 0.640\\    \cline{2-7}
\multicolumn{1}{|l|}{}                                   & Head 1 	& 0.11 $\pm$ \textit{0.06} & 0.15 $\pm$ \textit{0.08} & \textbf{0.21} $\pm$ \textit{0.16}& NOUN: 0.230& PER: \textbf{0.780}\\    \cline{2-7}
\multicolumn{1}{|l|}{}                                   & Head 2 	& \textbf{0.25} $\pm$ \textit{0.09} & \textbf{0.26} $\pm$ \textit{0.13} & 0.1 $\pm$ \textit{0.1}& NOUN: \textbf{0.560}& PER: 0.610\\    \cline{2-7}
\multicolumn{1}{|l|}{}                                   & Head 3 	& 0.09 $\pm$ \textit{0.07} & 0.12 $\pm$ \textit{0.13} & 0.13 $\pm$ \textit{0.13}& NOUN: 0.290& PER: 0.680\\    \cline{2-7}
\multicolumn{1}{|l|}{}                                   & Head 4 	& 0.18 $\pm$ \textit{0.06} & 0.18 $\pm$ \textit{0.09} & 0.11 $\pm$ \textit{0.11}& NOUN: \textbf{0.480}& PER: \textbf{0.830}\\    \cline{2-7}
\multicolumn{1}{|l|}{}                                   & Head 5 	& 0.14 $\pm$ \textit{0.08} & 0.16 $\pm$ \textit{0.09} & 0.1 $\pm$ \textit{0.09}& NOUN: 0.390& PER: 0.590\\    \cline{2-7}
\multicolumn{1}{|l|}{}                                   & Head 6 	& 0.06 $\pm$ \textit{0.03} & 0.12 $\pm$ \textit{0.07} & \textbf{0.22} $\pm$ \textit{0.19}& NOUN: 0.330& PER: 0.720\\    \cline{2-7}
\multicolumn{1}{|l|}{}                                   & Head 7 	& 0.12 $\pm$ \textit{0.07} & 0.15 $\pm$ \textit{0.11} & 0.09 $\pm$ \textit{0.1}& NOUN: 0.300& PER: 0.460\\    \hline \hline
\multicolumn{1}{|l|}{\multirow{8}{*}{Layer 3 }}  	 & Head 0 	& 0.07 $\pm$ \textit{0.04} & 0.15 $\pm$ \textit{0.08} & 0.2 $\pm$ \textit{0.17}& NOUN: 0.350& PER: 0.690\\    \cline{2-7}
\multicolumn{1}{|l|}{}                                   & Head 1 	& 0.17 $\pm$ \textit{0.12} & 0.09 $\pm$ \textit{0.05} & 0.11 $\pm$ \textit{0.11}& PUNC: 0.230& PER: \textbf{0.760}\\    \cline{2-7}
\multicolumn{1}{|l|}{}                                   & Head 2 	& 0.11 $\pm$ \textit{0.06} & 0.12 $\pm$ \textit{0.09} & 0.2 $\pm$ \textit{0.18}& NOUN: 0.420& PER: 0.620\\    \cline{2-7}
\multicolumn{1}{|l|}{}                                   & Head 3 	& \textbf{0.19} $\pm$ \textit{0.18} & 0.2 $\pm$ \textit{0.25} & 0.16 $\pm$ \textit{0.16}& NOUN: 0.350& ORG: 0.540\\    \cline{2-7}
\multicolumn{1}{|l|}{}                                   & Head 4 	& 0.1 $\pm$ \textit{0.06} & 0.14 $\pm$ \textit{0.09} & 0.09 $\pm$ \textit{0.08}& NOUN: 0.270& PER: 0.670\\    \cline{2-7}
\multicolumn{1}{|l|}{}                                   & Head 5 	& 0.11 $\pm$ \textit{0.06} & 0.14 $\pm$ \textit{0.06} & 0.16 $\pm$ \textit{0.14}& NOUN: 0.300& PER: 0.420\\    \cline{2-7}
\multicolumn{1}{|l|}{}                                   & Head 6 	& 0.16 $\pm$ \textit{0.07} & 0.18 $\pm$ \textit{0.1} & 0.1 $\pm$ \textit{0.1}& NOUN: \textbf{0.490}& PER: 0.680\\    \cline{2-7}
\multicolumn{1}{|l|}{}                                   & Head 7 	& 0.07 $\pm$ \textit{0.04} & 0.13 $\pm$ \textit{0.08} & 0.19 $\pm$ \textit{0.17}& NOUN: 0.360& PER: 0.750\\    \hline
\end{tabular}
\end{table*}

\begin{table}[]
\caption{Metric scores for the decoder cross attention of model 2. \#1 POS and \#1 NE show the most attended POS tag or named entity for that attention head along with its ratio compared to the other tags. For each column, the three heads with highest scores are boldfaced.}
\label{tab:model2_crossattn}
\begin{tabular}{ll|l|l|l|l|l|}
\cline{3-7}
                                                         &              & POS-KL & NEP & NER-KL & \#1 POS & \#1 NE \\ \hline
\multicolumn{1}{|l|}{\multirow{8}{*}{Layer 0 }}  	     & Head 0 	& 0.04 $\pm$ \textit{0.03} & 0.14 $\pm$ \textit{0.07} & 0.04 $\pm$ \textit{0.05}& NOUN: 0.320& PER: 0.480\\    \cline{2-7}
\multicolumn{1}{|l|}{}                                   & Head 1 	& 0.05 $\pm$ \textit{0.03} & 0.18 $\pm$ \textit{0.09} & 0.06 $\pm$ \textit{0.06}& NOUN: 0.370& PER: 0.580\\    \cline{2-7}
\multicolumn{1}{|l|}{}                                   & Head 2 	& 0.06 $\pm$ \textit{0.03} & 0.13 $\pm$ \textit{0.08} & \textbf{0.24} $\pm$ \textit{0.2}& NOUN: 0.310& PER: 0.560\\    \cline{2-7}
\multicolumn{1}{|l|}{}                                   & Head 3 	& 0.04 $\pm$ \textit{0.02} & 0.14 $\pm$ \textit{0.06} & 0.04 $\pm$ \textit{0.04}& NOUN: 0.350& PER: 0.490\\    \cline{2-7}
\multicolumn{1}{|l|}{}                                   & Head 4 	& 0.06 $\pm$ \textit{0.03} & 0.11 $\pm$ \textit{0.07} & 0.2 $\pm$ \textit{0.17}& NOUN: 0.280& ORG: 0.490\\    \cline{2-7}
\multicolumn{1}{|l|}{}                                   & Head 5 	& 0.05 $\pm$ \textit{0.03} & 0.19 $\pm$ \textit{0.09} & 0.08 $\pm$ \textit{0.07}& NOUN: 0.380& PER: 0.580\\    \cline{2-7}
\multicolumn{1}{|l|}{}                                   & Head 6 	& 0.17 $\pm$ \textit{0.05} & 0.14 $\pm$ \textit{0.08} & 0.06 $\pm$ \textit{0.05}& NOUN: 0.290& PER: 0.690\\    \cline{2-7}
\multicolumn{1}{|l|}{}                                   & Head 7 	& 0.1 $\pm$ \textit{0.04} & 0.17 $\pm$ \textit{0.07} & 0.06 $\pm$ \textit{0.06}& NOUN: \textbf{0.410}& PER: 0.520\\    \hline \hline
\multicolumn{1}{|l|}{\multirow{8}{*}{Layer 1 }}  	     & Head 0 	& 0.18 $\pm$ \textit{0.07} & 0.16 $\pm$ \textit{0.07} & 0.09 $\pm$ \textit{0.08}& PUNC: 0.290& PER: 0.560\\    \cline{2-7}
\multicolumn{1}{|l|}{}                                   & Head 1 	& 0.14 $\pm$ \textit{0.07} & 0.16 $\pm$ \textit{0.08} & 0.08 $\pm$ \textit{0.07}& NOUN: 0.370& PER: 0.620\\    \cline{2-7}
\multicolumn{1}{|l|}{}                                   & Head 2 	& 0.09 $\pm$ \textit{0.04} & 0.17 $\pm$ \textit{0.09} & 0.13 $\pm$ \textit{0.12}& NOUN: 0.400& PER: 0.600\\    \cline{2-7}
\multicolumn{1}{|l|}{}                                   & Head 3 	& 0.1 $\pm$ \textit{0.05} & 0.19 $\pm$ \textit{0.07} & 0.11 $\pm$ \textit{0.1}& NOUN: 0.310& PER: 0.440\\    \cline{2-7}
\multicolumn{1}{|l|}{}                                   & Head 4 	& 0.06 $\pm$ \textit{0.03} & 0.11 $\pm$ \textit{0.06} & 0.06 $\pm$ \textit{0.06}& NOUN: 0.360& PER: 0.620\\    \cline{2-7}
\multicolumn{1}{|l|}{}                                   & Head 5 	& 0.17 $\pm$ \textit{0.07} & 0.17 $\pm$ \textit{0.1} & 0.1 $\pm$ \textit{0.1}& NOUN: 0.370& PER: 0.710\\    \cline{2-7}
\multicolumn{1}{|l|}{}                                   & Head 6 	& \textbf{0.19} $\pm$ \textit{0.08} & 0.13 $\pm$ \textit{0.06} & 0.1 $\pm$ \textit{0.1}& VERB: 0.340& PER: 0.670\\    \cline{2-7}
\multicolumn{1}{|l|}{}                                   & Head 7 	& \textbf{0.22} $\pm$ \textit{0.1} & 0.17 $\pm$ \textit{0.08} & 0.1 $\pm$ \textit{0.1}& VERB: 0.320& PER: \textbf{0.740}\\    \hline \hline
\multicolumn{1}{|l|}{\multirow{8}{*}{Layer 2 }}  	     & Head 0 	& 0.08 $\pm$ \textit{0.03} & 0.18 $\pm$ \textit{0.09} & 0.1 $\pm$ \textit{0.09}& NOUN: 0.400& ORG: 0.500\\    \cline{2-7}
\multicolumn{1}{|l|}{}                                   & Head 1 	& 0.05 $\pm$ \textit{0.03} & 0.1 $\pm$ \textit{0.05} & 0.07 $\pm$ \textit{0.07}& NOUN: 0.300& PER: 0.670\\    \cline{2-7}
\multicolumn{1}{|l|}{}                                   & Head 2 	& 0.13 $\pm$ \textit{0.07} & 0.13 $\pm$ \textit{0.08} & 0.08 $\pm$ \textit{0.08}& NOUN: 0.310& PER: \textbf{0.830}\\    \cline{2-7}
\multicolumn{1}{|l|}{}                                   & Head 3 	& 0.05 $\pm$ \textit{0.02} & 0.16 $\pm$ \textit{0.08} & 0.08 $\pm$ \textit{0.08}& NOUN: 0.290& PER: 0.550\\    \cline{2-7}
\multicolumn{1}{|l|}{}                                   & Head 4 	& 0.06 $\pm$ \textit{0.03} & 0.12 $\pm$ \textit{0.07} & \textbf{0.21} $\pm$ \textit{0.18}& NOUN: 0.300& PER: 0.520\\    \cline{2-7}
\multicolumn{1}{|l|}{}                                   & Head 5 	& 0.14 $\pm$ \textit{0.05} & 0.16 $\pm$ \textit{0.08} & 0.08 $\pm$ \textit{0.08}& NOUN: 0.390& PER: 0.610\\    \cline{2-7}
\multicolumn{1}{|l|}{}                                   & Head 6 	& \textbf{0.24} $\pm$ \textit{0.09} & \textbf{0.23} $\pm$ \textit{0.12} & 0.07 $\pm$ \textit{0.08}& NOUN: 0.430& PER: 0.690\\    \cline{2-7}
\multicolumn{1}{|l|}{}                                   & Head 7 	& 0.09 $\pm$ \textit{0.04} & 0.1 $\pm$ \textit{0.06} & \textbf{0.21} $\pm$ \textit{0.19}& NOUN: 0.340& PER: \textbf{0.760}\\    \hline \hline
\multicolumn{1}{|l|}{\multirow{8}{*}{Layer 3 }}     	 & Head 0 	& 0.08 $\pm$ \textit{0.05} & 0.16 $\pm$ \textit{0.09} & 0.2 $\pm$ \textit{0.17}& NOUN: 0.360& PER: 0.470\\    \cline{2-7}
\multicolumn{1}{|l|}{}                                   & Head 1 	& 0.11 $\pm$ \textit{0.05} & \textbf{0.21} $\pm$ \textit{0.09} & 0.12 $\pm$ \textit{0.12}& NOUN: \textbf{\textbf{0.410}}& PER: 0.680\\    \cline{2-7}
\multicolumn{1}{|l|}{}                                   & Head 2 	& 0.12 $\pm$ \textit{0.12} & \textbf{0.24} $\pm$ \textit{0.18} & 0.11 $\pm$ \textit{0.12}& NOUN: \textbf{0.420}& PER: 0.490\\    \cline{2-7}
\multicolumn{1}{|l|}{}                                   & Head 3 	& 0.15 $\pm$ \textit{0.07} & 0.17 $\pm$ \textit{0.09} & 0.16 $\pm$ \textit{0.15}& NOUN: 0.400& PER: 0.490\\    \cline{2-7}
\multicolumn{1}{|l|}{}                                   & Head 4 	& 0.07 $\pm$ \textit{0.03} & 0.11 $\pm$ \textit{0.06} & \textbf{0.21} $\pm$ \textit{0.18}& NOUN: 0.370& PER: 0.570\\    \cline{2-7}
\multicolumn{1}{|l|}{}                                   & Head 5 	& 0.08 $\pm$ \textit{0.09} & 0.12 $\pm$ \textit{0.08} & 0.1 $\pm$ \textit{0.11}& NOUN: 0.290& PER: 0.510\\    \cline{2-7}
\multicolumn{1}{|l|}{}                                   & Head 6 	& 0.1 $\pm$ \textit{0.05} & 0.13 $\pm$ \textit{0.08} & 0.11 $\pm$ \textit{0.11}& NOUN: 0.360& PER: 0.700\\    \cline{2-7}
\multicolumn{1}{|l|}{}                                   & Head 7 	& 0.07 $\pm$ \textit{0.03} & 0.12 $\pm$ \textit{0.07} & 0.13 $\pm$ \textit{0.12}& NOUN: 0.330& PER: 0.730\\    \hline
\end{tabular}
\end{table}
\end{document}